\documentclass{article}

\usepackage{microtype}
\usepackage{xcolor}
\usepackage{graphicx}
\usepackage{subfigure}
\usepackage{booktabs} 

\usepackage{hyperref}

\usepackage{tcolorbox}
\usepackage{listings}

\usepackage{booktabs}
\usepackage{multirow}
\usepackage{array}
\usepackage{subcaption}

\usepackage{subcaption}

\usepackage[accepted]{icml2025}

\usepackage{amsmath}
\usepackage{amssymb}
\usepackage{mathtools}
\usepackage{amsthm}

\usepackage{multirow}

\usepackage[capitalize,noabbrev]{cleveref}

\theoremstyle{plain}

\theoremstyle{definition}

\theoremstyle{remark}

\usepackage[textsize=tiny]{todonotes}

\begin{document}

\twocolumn[
\icmltitle{GTM: Simulating the World of Tools for AI Agents}




\begin{icmlauthorlist}

\icmlauthor{Zhenzhen Ren}{fudan,zga}
\icmlauthor{Xinpeng Zhang}{fudan}
\icmlauthor{Zhenxing Qian}{fudan}
\icmlauthor{Yan Gao}{zga}
\icmlauthor{Yu Shi}{zga}
\icmlauthor{Shuxin Zheng}{zga}
\icmlauthor{Jiyan He}{zga}

\end{icmlauthorlist}

\icmlaffiliation{zga}{Zhongguancun Academy, Beijing, China}
\icmlaffiliation{fudan}{School of Computer Science, Fudan University, Shanghai, China}

\icmlcorrespondingauthor{Jiyan He}{hejiyan@zgci.ac.cn}


\vskip 0.3in
]



\printAffiliationsAndNotice{\icmlEqualContribution} 

\begin{abstract}

The integration of external tools is pivotal for empowering Large Language Model (LLM) agents with real-world capabilities. 
However, training these agents through direct, continuous interaction with diverse tools is often prohibitively expensive, slow, and introduces additional development and maintenance overhead.
To address this challenge, we introduce the Generalist Tool Model (GTM), a 1.5-billion-parameter model that learns to act as a universal tool simulator. 
With only prompt-level configuration, GTM accesses tool functionalities along with input arguments and generates outputs that faithfully mimic real tool execution, providing a fast and cost-effective solution that eliminates development overhead.
To build GTM, we propose the Context-Aware Response Generation (CARG) pipeline, which synthesizes comprehensive training data covering over 20,000 tools across 300 domains including physics, medicine, robotics, and finance.
Through this pipeline, GTM learns to produce not only syntactically correct outputs but also logically coherent and contextually appropriate responses.
Experiments demonstrate that GTM produces high-quality outputs with strong consistency and reliability. Besides when used in real reinforcement learning scenarios for agent training, GTM exhibits significantly faster simulation speed compared to real tools while maintaining comparable output quality, along with remarkable generalization and domain adaptability. 
Our results establish GTM as a foundational component for developing future AI agents, enabling efficient and scalable training of tool-augmented systems.


\end{abstract}







\section{Introduction}

Large Language Models (LLMs) have demonstrated exceptional capabilities in natural language understanding, reasoning, and generation \cite{chkirbene2024large}. 
However, they operate primarily within the domain of text processing and lack the ability to directly access external systems or perform actions in the physical world \cite{schick2023toolformer,qin2023toolllm}. 
This disconnect between language comprehension and real-world interaction inherently limits what LLMs can achieve on their own. 
Tools, which can be regarded as functions that take specific inputs and produce corresponding outputs through predefined 
operations, offer a promising solution by bridging this gap \cite{masterman2024landscape,yao2023react}. 
By learning to invoke appropriate tools, LLM-based agents can extend their functionality beyond text generation to carry out concrete tasks such as executing code, querying databases, or controlling robotic systems \cite{wang2024executable,luo2024integration,zhuang2023toolqa}.

\begin{figure*}[htbp] 
    \centering 
    
    \includegraphics[width=1\textwidth]{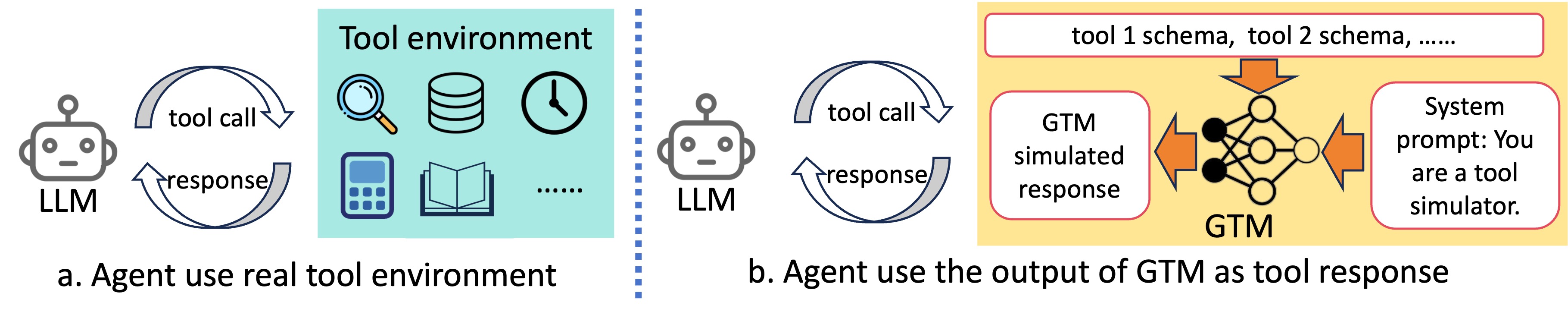} 
    \caption{Comparison between the process of use real tool and use GTM simulated tools. a. Real tool environment need various tools. b. With only prompt-level modification, GTM can simulate various tools, thus providing a more generous choice for agent tool learning.} 
    \label{fig:teaser_fig} 
\end{figure*}

Previous work has explored teaching agents to use tools by training them to produce correctly formatted calls, interpret tool outputs, and incorporate the results into subsequent reasoning \cite{tang2023toolalpaca,wang2024metatool,qin2023toolllm}. 
Tool learning has been approached through two main methods. The first is supervised fine-tuning (SFT), which trains agents on curated datasets of tool-use examples \cite{qin2023toolllm,tang2023toolalpaca,ye2024tooleyes,liu2024apigen,chakraborty2025t1,wu2024seal,basu2024api}. 
Though effective for learning basic patterns, SFT faces challenges in generalization and lacks the ability to explore and adapt to new situations \cite{qian2025toolrl}. 
The second approach is reinforcement learning (RL), which has proven effective in enhancing LLMs' reasoning and generalization capabilities \cite{guo2025deepseek,xie2025logic}.
In RL, agents learn through interaction with an environment, optimizing their behavior to maximize cumulative rewards \cite{cao2024survey,wang2024reinforcement}. 
Within this framework, tools serve dual purposes: they function as actions that agents can invoke to gather information or perform computations, and some tools can also act as evaluators to help assess the effectiveness of the agent's outputs 
\cite{li2025torl,qian2025toolrl,feng2025retool,yu2024steptool,singh2025agentic}. 

Despite the success of RL in enabling agents to discover effective tool-use strategies and adapt to diverse scenarios, integrating tools into RL training processes presents significant practical challenges. First, many tools are external API calls that introduce considerable latency, dramatically slowing down the training process. For instance, web search APIs like Jina Search typically require 2-3 seconds per query and impose rate limits (e.g., 100 queries per minute), which can bottleneck RL algorithms that require millions of interactions \cite{zheng2025deepresearcher,chen2025reinforcement,jin2025search,song2025r1,li2025webthinker}.
Second, tool invocations can be prohibitively expensive, especially for compute-intensive services, making large-scale RL training with millions of tool calls economically infeasible \cite{sun2025zerosearch,song2025help}.
Third, integrating external tools introduces significant engineering overhead: developing robust interfaces, handling diverse response formats, debugging integration issues, and maintaining compatibility as APIs evolve all require substantial effort that diverts resources from core algorithm development \cite{sun2025zerosearch,luo2025agent}.
These practical constraints pose fundamental obstacles to scaling tool-augmented RL systems.

To address these challenges, we propose the Generalist Tool Model (GTM), a universal tool simulator that generates outputs mimicking real tool responses instead of invoking actual APIs during training.
As shown in Figure ~\ref{fig:teaser_fig}, with only prompt-level modification, GTM can simulate various tools.
GTM offers several key advantages that directly address the aforementioned challenges.
First, as a 1.5-billion-parameter model, GTM enables the use of existing high-speed inference frameworks (e.g. LMDeploy \cite{zhang2025efficient}), allowing batch tool calls to achieve significantly higher throughput and lower latency compared to traditional API requests, effectively solving the speed bottleneck.
Second, it transforms expensive per-call API costs into predictable GTM inference costs, making large-scale tool-augmented training economically viable.
Third, GTM eliminates the engineering overhead of tool integration by providing a unified interface: developers no longer need to handle diverse APIs, debug network issues, or maintain multiple tool integrations, as GTM provides consistent outputs across all simulated tools.

The design philosophy of GTM focuses on simulating tools that progress from format correctness, to logical coherence, to context consistency.
the Context Awareness Response Generation (CARG) algorithm, which constructs high-quality data for training GTM. 
CARG consists of two stages: a generation stage and a validation stage. CARG can produce both single-turn tool responses and multi-turn contextually consistent tool responses.
Experiments across three scenarios, including seen tools, unseen tools, and domain-specific tools, demonstrate that GTM simulates tools significantly faster while exhibiting strong generalization and domain adaptability, achieving performance comparable to real tools.
Our contributions are threefold:
\begin{itemize}
\item We first propose the Generalist Tool Model (GTM), a foundational component of RL training that can simulate diverse tool behaviors without accessing real tool implementations, enabling efficient development of tool use agents.
\item We propose Context Awareness Response Generation (CARG) pipeline, which teaches GTM to learn not only format correctness, but also logical coherence and context consistency in tool responses.
\item We demonstrate that when used in real RL training processes, GTM exhibits faster simulation speed, quality comparable to real tools, and strong generalization and domain adaptability.
\end{itemize}

\section{Related Work}


\subsection{Agent Tool Learning}
Tool use has emerged as a critical capability for LLM-based agents to overcome their inherent limitations in perceiving and manipulating the external world. The core challenge lies in helping models understand the "meta-properties" of tools, namely their essential, task-transferable characteristics such as causal relationships and operational constraints \cite{tang2023toolalpaca,wang2024metatool,qin2023toolllm}. Recent research has explored two primary paradigms: supervised fine-tuning (SFT) and reinforcement learning (RL).

SFT approaches construct datasets following a two-stage pipeline: defining tool APIs with clear specifications, then generating conversational trajectories demonstrating appropriate tool invocations. 
While methods using real-world tools \cite{qin2023toolllm,jiang2025medagentbench,tang2023toolalpaca,ye2024tooleyes} or synthetic data \cite{tao2024harnessing,wu2024seal,liu2024apigen,huang2025advancing,wang2024toolflow,shi2025tool,liu2024toolace} have shown promise results, SFT fundamentally struggles with generalization to novel situations \cite{qian2025toolrl}.
RL methods address these limitations by enabling learning through environmental interaction. 
DeepResearcher \cite{zheng2025deepresearcher} and WebAgent-R1 \cite{wei2025webagent} exemplify this approach, embedding tools within iterative workflows where agents learn optimal strategies through trial-and-error. 
However, practical deployment faces significant barriers: external API calls drastically slow training, tool responses are unstable, and per-request pricing makes exploration prohibitively expensive.

\subsection{Optimizing Agentic RL Training}
Agentic RL training refers to the process of using reinforcement learning to optimize LLM-based agents that interact with complex environments through actions, observations, and rewards. 
Recent work has focused on decoupling various components of the RL training pipeline to improve efficiency and scalability. 
One line of research develops general-purpose reward models that can evaluate agent behaviors across diverse tasks. Recent Generalist Reward Models (GRMs) like DeepSeek-GRM \cite{liu2025inference} leverage generative reward modeling with inference-time scaling, while \cite{li2025generalist} discovers latent reward signals within pre-trained LLMs. While GRMs focus on evaluating agent behaviors, our GTM takes a complementary approach by simulating tool responses to decouple agents from external dependencies during training.
Another direction separates RL inference from training, allowing asynchronous policy updates and distributed execution\cite{luo2025agent}. 
Additionally, several works address the scalability of RL training environments\cite{fang2025towards}.
A notable attempt at tool decoupling is ZeroSearch\cite{sun2025zerosearch}, which successfully eliminates API dependencies by replacing web search with a model-based simulator. However, this approach is specifically designed for web search and cannot be extended to other tools in the agent ecosystem.
For the broader landscape of tool-augmented agents, these fundamental challenges persist: external API calls slow down training, tool responses are unstable, and API costs during exploration remain prohibitively high.
Our work aims to address these challenges by decoupling tools from the agent RL training process through a generalist tool model.




\section{Problem Formulation}
\subsection{Usage Scenario Modeling}
In a typical RL training pipeline for tool-learning agents, there are several key components: an actor model (the agent being trained), a reference model that constrains the actor to prevent deviation from initial objectives, tools that provide external capabilities, and a reward function or model that evaluates the agent's performance. 
During each training iteration, the actor model generates tool invocations in specified formats to execute queries, access external services, and so on. 
It then interpret the tool outputs to either produce final results or determine the next tool call in a multi-step process.
Once the actor model produces a final output, a reward module, which can be either a model or a tool itself, evaluates the quality of the result and provides feedback signals.
GTM seamlessly integrates into this pipeline by simulating tool outputs, effectively serving as either the tools themselves or as part of the reward function.
\subsection{Design Goals of GTM}
The core design philosophy of GTM is to decouple the tool component from tool-learning agent training, enabling efficient tool learning by simulating tool outputs instead of invoking real APIs. Tool learning typically encompasses several objectives: generating properly formatted calls, interpreting tool outputs, and incorporating the results into subsequent reasoning. To support these learning objectives effectively, GTM is designed with the following goals:
\begin{itemize}
    \item [1)] \textbf{Correctness}: GTM should ensure that tool invocations follow correct syntax and structure, helping agents learn proper API calling conventions and parameter formatting requirements.
    \item [2)] \textbf{Generalization}: GTM should simulate a wide range of tools across diverse domains and functionalities, eliminating the need for tool learning in most cases.
    \item[3)] \textbf{Consistency}: Consistency means not only maintaining semantic coherence between inputs and outputs but also preserving contextual continuity across multi-turn interactions.
    \item [4)] \textbf{Helpfulness and Usefulness}: Helpfulness refers to its capability to generate appropriate error messages for incorrect inputs, while usefulness refers to GTM's ability to provide meaningful feedback that effectively guides the agent's learning during the real RL training process.
\end{itemize}


%



\begin{figure}[h]
\begin{tcolorbox}[
    colback=blue!5!white,
    colframe=blue!75!black,
    title={\textbf{Unified Tool Template}},
    fonttitle=\bfseries\small,
    boxrule=0.5pt,
    width=\columnwidth
]
\begin{lstlisting}[basicstyle=\scriptsize\ttfamily, escapeinside={(*}{*)}]
{
  (*\textcolor{blue}{"api\_name"}*): string,
  (*\textcolor{blue}{"api\_description"}*): string,
  (*\textcolor{blue}{"field"}*): string,
  (*\textcolor{blue}{"parameters"}*): {
    <param_name>: {
      (*\textcolor{red}{"type"}*): string,
      (*\textcolor{red}{"description"}*): string
    }
  },
  (*\textcolor{blue}{"required"}*): [param_name, ...],
  (*\textcolor{blue}{"responses"}*): {
    <response_field>: {
      (*\textcolor{red}{"type"}*): string,
      (*\textcolor{red}{"description"}*): string
    }
  }
}
\end{lstlisting}
\end{tcolorbox}
\caption{Unified Tool Template Structure}
\label{fig:tool-template}
\end{figure}

\section{Method}

GTM is designed to simulate diverse tools at an affordable computational cost. 
We selected Qwen2.5-1.5B as our base model, striking a balance between capability and cost. 
GTM training consists of three key steps: 
(1) Tool generation, where we create a comprehensive collection of tool specifications;
(2) Context Awareness Response Generation (CARG) pipeline, which generates contextually coherent responses by understanding tool-specific context and maintaining continuity across multiple tool invocations in multi-turn interactions.
(3) Training, where we fine-tune the model to achieve our design goals.



\subsection{Tool Generation}

We build a comprehensive and diverse tool collection by drawing inspiration from the pipeline of Seal-Tools, a framework originally designed for generating large-scale tool-calling dialogue datasets. However, unlike Seal-Tools which focuses on creating natural tool-calling conversations, we target the generation of varied tool schemas. Additionally, we extract tool specifications from two other large-scale tool learning datasets: ToolEyes and APIGen. To ensure consistency across our tool collection, we design a unified tool format template as shown in Figure~\ref{fig:tool-template}. Our tool generation process follows three systematic steps: (1) Field/subfield generation, (2) Tool generation and validation, and (3) Integration of external tools, where we extract, standardize, and deduplicate tools from ToolEyes and APIGen.

\begin{figure}[htbp]
    \centering
    \includegraphics[width=\columnwidth, trim=2cm 3.0cm 2cm 2.7cm, clip]{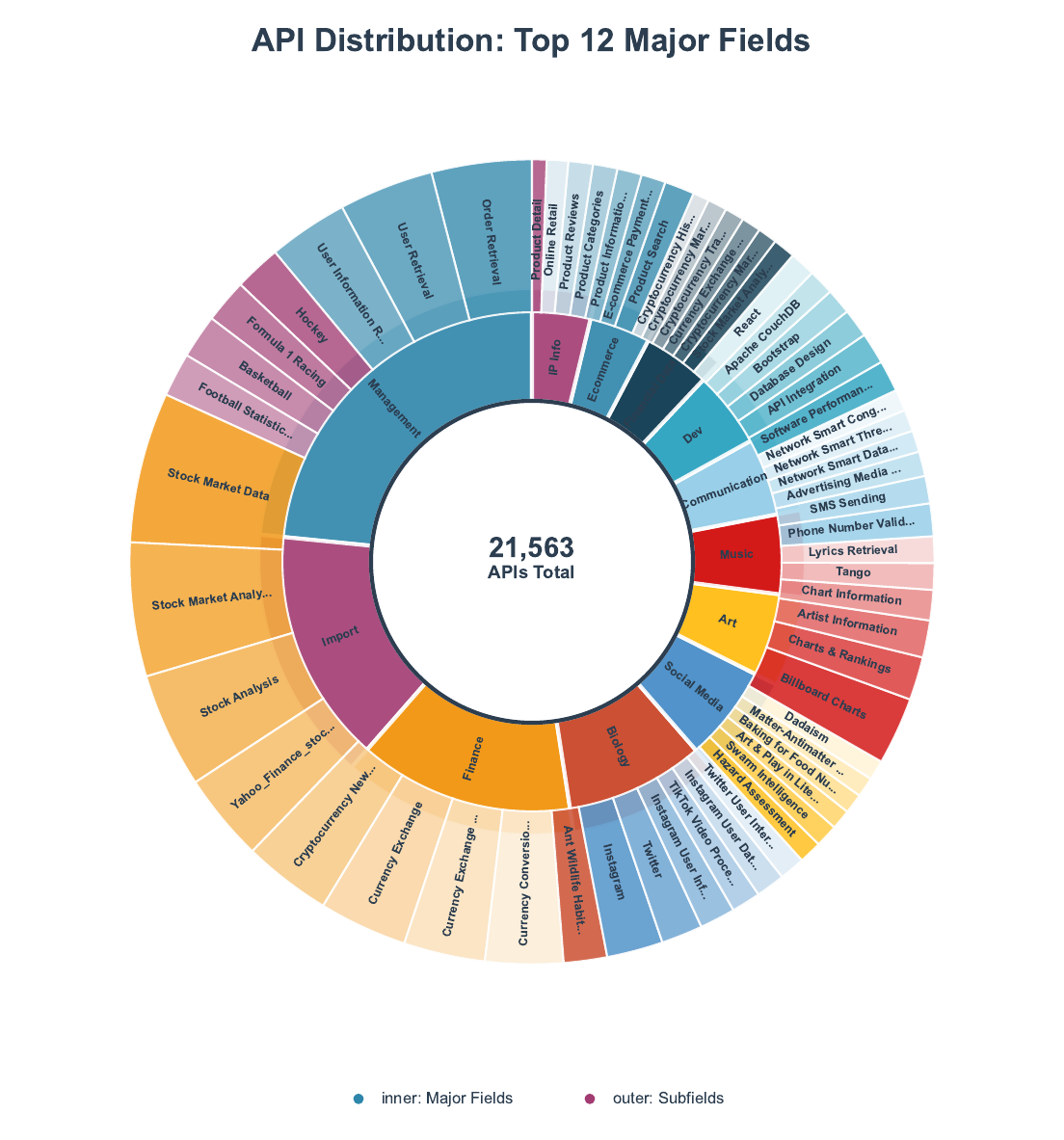}
    \caption{Top 12 field overview. Inner: field, Outer: subfield}
    \label{fig:12_field}
\end{figure}

First, we build our tool taxonomy through iterative expansion. Starting from seed fields $\mathcal{F}_0$ (e.g., "finance", "healthcare", "education"), we iteratively generate new fields by prompting an LLM with randomly sampled exemplar pairs from existing fields. For instance, given "finance" and "healthcare" as exemplars, the LLM might generate "insurance" as a new field. This process continues until we reach sufficient coverage across diverse domains. For each field $f \in \mathcal{F}$, we then generate specialized subfields $\mathcal{S}_f$. For example, under "finance", we create subfields like "stock trading", "banking", and "cryptocurrency". This creates a two-level hierarchy that organizes our tool collection systematically.

Second, for each field-subfield pair $(f, s)$ where $f \in \mathcal{F}$ and $s \in \mathcal{S}_f$, we generate a set of tools $\mathcal{T}_{f,s}$. Each tool follows a strict schema that includes tool name, description, parameters with types, required parameters, and expected response format. To ensure quality, we perform two critical checks. We first validate that all required parameters actually exist in the parameter list, which is a common error in LLM-generated content. We then deduplicate tools by computing cosine similarity between tool names, removing those with similarity above 0.8 to avoid redundancy. This process yields a high-quality collection of diverse, well-specified tools.

Finally, to maximize coverage, we integrate tools from existing repositories including ToolEyes and APIGen. We extract tool specifications from these datasets and transform them into our unified format. For tools with missing information such as parameter types or response formats, we use LLMs to complete the specifications based on tool descriptions. We then apply the same validation and deduplication process as above, ensuring consistency across our entire tool collection. The final tool repository $\mathcal{T}$ contains over 20,000 unique tools spanning diverse domains and functionalities. As shown in Figure~\ref{fig:12_field}, our collection covers a wide range of fields from finance and communication to art and social media, with each field containing multiple specialized subfields that contain diverse tool patterns.

\subsection{Context Awareness Response Generation Strategy}

To generate tool input and output for a given API call, our goal extends beyond producing syntactically correctness. We aim to create responses that are semantically meaningful, contextually appropriate, and consistent across multi-turn interactions. The Context Awareness Response Generation (CARG) pipeline employs a two-stage "generate-validate" architecture to generate data in 3 scenario: single-turn input-output generation; multi-turn contextual generation; and error generation. 

\subsubsection{Single-turn input-output generation.} 
In the single-turn setting, CARG fucus on generating data that demonstrate understanding of the tool's functional semantics, maintain logical relationships between parameters, and produce outputs that meaningfully correspond to the given inputs within realistic usage scenarios.

\textbf{a. Generation stage}: For each tool $\tau$ with specification $\mathcal{T}_\tau = (d_\tau, \mathcal{P}, \mathcal{R})$, in which $d_\tau$ stands for tool description, $\mathcal{P}$ stands for parameters' information and $\mathcal{R}$ stands for response information, we prompt an LLM $\mathcal{M}$ to synthesize contextually-aware examples by understanding the tool's purpose and domain. The model generates:

\begin{equation}
\{(x, y)\} = \mathcal{M}(\mathcal{T}_\tau, \mathcal{C}_{\text{domain}}, \text{prompt}_{\text{gen}})    
\label{eqa:gen}
\end{equation}

where $\mathcal{C}_{\text{domain}}$ represents domain-specific context (e.g., geographical knowledge for location APIs), $x = \{p_1: v_1, ..., p_k: v_k\}$ contains semantically meaningful parameter values, and $y$ is an output that logically follows from the input context.

\textbf{b. Validation stage}: Each generated pair undergoes three levels of verification to ensure quality and contextual appropriateness.
Tools with persistently low validation rates undergo multiple generation attempts with refined prompts, ensuring balanced
representation and high contextual quality across all tools
in the final dataset.
\begin{equation}
V(x, y) = \left\{
\begin{aligned}
&V_{\text{format}}(x, y) \\
&V_{\text{logic}}(x, y) = \mathcal{M}(x, y, \text{prompt}_{\text{logic}}) \\
&V_{\text{sem}}(x, y) = \mathcal{M}(\mathcal{T}_\tau, x, y, \text{prompt}_{\text{sem}})
\end{aligned}
\right.
\label{eqa:validation}
\end{equation}
where $V_{\text{format}}$ checks parameter types and required fields, $V_{\text{logic}}$ detects parameter contradictions, and $V_{\text{sem}}$ verifies input-output coherence. Each validation function $V_i(x, y) \in \{\text{pass}, \text{fail}\}$. Only pairs that pass all three validation levels (i.e., $V_{\text{format}}(x, y) = V_{\text{logic}}(x, y) = V_{\text{sem}}(x, y) = \text{pass}$) proceed to the filtering stage.


\subsubsection{Multi-turn contextual generation.}
In the multi-turn setting, CARG extends the single-turn approach to generate dialogue sequences that demonstrate progressive context building, cross-turn information dependencies, and coherent tool invocations that naturally emerge from conversational flow.

\textbf{a. Generation stage}: The multi-turn generation process consists of two key steps: API semantic association; Progressive dialogue generation.

\textit{API semantic association}: We first encode all APIs into high-dimensional vectors using SentenceTransformer: $\mathbf{e}_{\tau_i} = \text{SentenceTransformer}(\text{concat}(\text{name}_{\tau_i}, \text{desc}_{\tau_i}, \text{domain}_{\tau_i}))$, where each API's representation includes its name, description, and domain. Starting from a randomly selected seed API, we employ greedy search to construct semantically coherent API groups:
\begin{equation}
\mathcal{G} = \{\tau_j : \cos(\mathbf{e}_{\tau_i}, \mathbf{e}_{\tau_j}) > \theta, \tau_i \in \mathcal{G}\}
\end{equation}
where $\theta$ ensures semantic relevance. The group coherence score $C(\mathcal{G}) = \frac{1}{|\mathcal{G}|^2}\sum_{i,j}\cos(\mathbf{e}_{\tau_i}, \mathbf{e}_{\tau_j})$ serves as a quality indicator.

\textit{Progressive dialogue generation}: We construct multi-turn dialogues by distributing selected APIs across $L$ turns, with the final API as the target tool call. Each turn $t$ is generated with accumulated context $\mathcal{C}_t = \bigcup_{i=1}^{t-1} \text{context\_updates}_i$, where context updates contain key information from previous turns. The generation follows:
\begin{equation}
\text{turn}_t = \mathcal{M}(\mathcal{T}_{\tau_t}, \mathcal{C}_t, \mathcal{H}_{t-1}, \text{prompt}_{\text{dialogue}})
\end{equation}
where $\mathcal{H}_{t-1}$ represents the history of the conversation. The final call to the tool $y = \mathcal{M}(\mathcal{H}_L, \mathcal{C}_L, \tau_{\text{target}})$ takes advantage of the complete dialog context to ensure coherent parameter generation.

\textbf{b. Validation stage}: Building upon the single-turn validation pipeline, we augment the validation process with dialog-specific checks:
\begin{equation}
V_{\text{multi}}(\mathcal{H}, y) = V(x_L, y) \land V_{\text{coherence}}(\mathcal{H}, y)
\end{equation}
where $V(x_L, y)$ applies the three-level validation from Equation~\ref{eqa:validation} to the final tool call, and $V_{\text{coherence}}(\mathcal{H}, y) = \mathcal{M}(\mathcal{H}, y, \text{prompt}_{\text{coherence}})$ additionally verifies that the tool invocation logically follows from the conversation history.

\subsubsection{Error generation.}
To enhance model robustness in handling incorrect API calls, CARG generates diverse error scenarios that simulate common mistakes paired with appropriate error messages.

\textbf{a. Generation stage}: We implement four specialized error generators to create realistic failure cases: (1) \textit{Type error generator} produces parameter values with incorrect data types; (2) \textit{Missing required parameter generator} omits mandatory parameters specified in the API definition; (3) \textit{Excess parameter generator} adds non-existent parameters to the API call; (4) \textit{Invalid value generator} generates semantically inappropriate values while maintaining correct types. For each error type, we transform a valid input from Equation~\ref{eqa:gen} and generate corresponding error messages:
\begin{equation}
(x_{err}, e_{msg}) = \mathcal{M}(\mathcal{T}_\tau, \text{error\_type}, x_{valid}, \text{prompt}_{err})
\end{equation}
where $x_{valid}$ is a valid input, $x_{err}$ is the erroneous version, and $e_{msg}$ is the generated error message explaining the specific issue.

\textbf{b. Validation stage}: Each error-message pair undergoes three-level verification: format validation, error existence validation, and error message quality assessment. The complete validation is formalized as:
\begin{equation}
V_{err}(x_{err}, e_{msg}) = V_{format} \land V_{exist} \land V_{quality}
\end{equation}
where each validation function verifies specific aspects of the error-message pair against the API definition $\mathcal{T}_\tau$. Only pairs passing all validation levels are retained.

\subsection{Training of GTM}

To maximize the generalization of GTM in tool-world simulation, we adopt the approach of fine-tuning open-source large language models (LLMs). This choice is motivated by the extensive knowledge accumulated by LLMs during their pre-training on massive datasets, which spans diverse domains, reasoning patterns, and factual information, all of which provide a solid foundation for modeling complex tool-world interactions. By leveraging such pre-trained LLMs as the base architecture, GTM can inherit inherent abilities to understand tool functionalities, predict user intentions, and simulate plausible interaction sequences, reducing the need for building from scratch in tool-world scenarios. Specifically, we utilized the CARG pipeline to generate data for fine-tuning the Qwen2.5-1.5B model. We selected such model parameters to balance performance and computational cost. Ultimately, we achieved the GTM-1.5B model.

\section{Experiments and Analysis}
This section outlines two main groups of experiments designed to evaluate the performance of GTM and its practical utility. The first group focuses on validating GTM's ability to generate not only format-correct outputs but also logically coherent and contextually consistent tool responses, which includes comparisons with common open-source models in three scenarios: single-turn dialog, multi-turn dialog, and inputs containing errors. The second group verifies the utility of GTM in real reinforcement learning (RL) processes, covering three practical RL scenarios: search, retrieval, and kernel optimization. Notably, the tools involved in the retrieval and kernel optimization scenarios have no similar counterparts in GTM's training dataset.

For the first group of experiments, we utilized GTM models fine-tuned from Qwen2.5-1.5B, with the API dataset split into training and validation sets. These GTM variants were compared against several common open-source models, including the Qwen2.5 series, InternLM2.5 series, and Llama3.1 series. To ensure objective evaluation, a Qwen2.5-72B model was employed as the judger to assess output quality across the three scenarios.

For the second group of experiments, we utilized GTM to simulate real-world tools across three distinct scenarios: search, retrieval, and kernel optimization. These scenarios were specifically chosen to demonstrate GTM's effectiveness under different conditions. The search scenario involves tools similar to those in our training dataset, while the retrieval scenario represents a completely novel tool type absent from the training data. The kernel optimization scenario was included to evaluate GTM's domain adaptation capabilities, as it represents a highly specialized field where tools are domain-specific and cannot be directly simulated by the base GTM model, thus requiring fine-tuning. For the search tool, we selected Jina's search API with a concurrency limit of 40 requests per minute. For the retrieval tool, we adopted the same configuration used in Search-R1. For both search and retrieval, we followed a consistent experimental protocol: during training, GTM simulates tool responses by generating outputs based on provided ground-truth answers, while during testing, all models interact with actual tools to ensure fair comparison. For kernel optimization, we utilized GTM to simulate a specialized tool that assesses code properties such as compilation errors and runtime performance.

\begin{table*}[htbp]
\centering
\caption{Model Evaluation Results (\%)}
\label{tab:model_evaluation}
\footnotesize
\setlength{\tabcolsep}{2.5pt} 
\begin{tabular}{l|cccccccccccccc|c}
\toprule
\multirow{2}{*}{Model} & \multicolumn{5}{c|}{Single-turn} & \multicolumn{6}{c|}{Multi-turn} & \multicolumn{3}{c|}{Error Detection} & \multirow{2}{*}{Avg} \\
\cmidrule{2-15}
 & Format & Logic & Sem & Comp & All & Format & Logic & Sem & Comp & Cons & All & Det & Help & All & \\
\midrule



Qwen2.5-0.5B-Instruct & 97.4 & 83.4 & 83.8 & 97.0 & 78.1 & 66.7 & 47.0 & 43.2 & 70.4 & 93.4 & 33.4 & 76.8 & 30.3 & 23.9 & \textbf{45.1} \\

Qwen2.5-1.5B-Instruct & 97.0 & 93.1 & 94.8 & 97.0 & 90.3 & 74.1 & 74.5 & 68.9 & 84.8 & 98.4 & 53.0 & 82.1 & 47.0 & 40.2 &  \textbf{61.2}\\

Qwen2.5-3B-Instruct & 95.3 & 94.7 & 94.9 & 98.7 & 89.6 & 80.0 & 85.4 & 83.4 & 93.9 & 99.2 & 65.8 & 81.2 & 47.0 & 40.7 &  \textbf{65.3} \\

Qwen2.5-7B-Instruct & 99.6 & 99.6 & 99.3 & 99.9 & 98.8 & 91.4 & 94.9 & 95.0 & 97.2 & 99.5 & 85.6 & 89.0 & 69.3 & 64.7 &  \textbf{83.0} \\

Qwen2.5-14B-Instruct & 99.3 & 99.6 & 99.9 & 99.9 & 98.8 & 88.0 & 96.4 & 96.7 & 98.6 & 99.7 & 84.0 & 85.2 & 81.6 & 74.6 & \textbf{85.8}\\

Llama-3.2-1B-Instruct & 69.6 & 58.1 & 51.6 & 80.3 & 45.6 & 38.3 & 30.7 & 21.9 & 45.8 & 95.3 & 13.4 & 80.1 & 23.2 & 20.6 &  \textbf{39.8} \\

Llama-3.2-3B-Instruct & 96.3 & 94.5 & 93.9 & 97.6 & 89.3 & 82.4 & 74.5 & 72.9 & 83.5 & 97.9 & 58.3 & 87.2 & 62.0 & 57.5 & \textbf{68.3} \\

InternLM2.5-1.8B & 56.0 & 8.7 & 5.0 & 33.0 & 3.0 & 34.8 & 19.3 & 12.4 & 24.8 & 94.8 & 5.2 & 91.0 & 25.9 & 24.2 & \textbf{10.8} \\

InternLM2.5-7B & 72.2 & 70.0 & 63.0 & 70.9 & 53.6 & 63.6 & 48.9 & 40.9 & 51.2 & 98.9 & 32.4 & 86.5 & 35.4 & 31.8 & \textbf{39.2} \\

InternLM2.5-20B & 86.4 & 92.6 & 92.8 & 96.4 & 81.8 & 79.8 & 80.4 & 78.7 & 83.3 & 99.6 & 68.5 & 86.5 & 65.9 & 58.1 & \textbf{69.4} \\
\midrule


GTM-1.5B & 99.2 & 97.3 & 97.6 & 98.0 & 95.5 & 97.2 & 90.1 & 94.0 & 98.5 & 99.0 & 86.7 & 95.3 & 87.5 & 86.1 & \textbf{89.4} \\


\bottomrule
\end{tabular}
\end{table*}

\subsection{Evaluation of GTM's Output Quality}.

Prior to validating GTM's utility in real RL processes, it is critical to first confirm that GTM satisfies our three core objectives: correctness in syntax and structure, consistency in maintaining semantic coherence and contextual continuity, and helpfulness in generating appropriate error messages.
To quantify these properties, we defined a set of evaluation metrics: Format (adherence to correct syntax and structure), Logic (absence of logical contradictions), Sem (semantic coherence between inputs and outputs), Comp (inclusion of all required fields), and Cons (contextual continuity across multi-turn interactions). For each scenario (single-turn, multi-turn, error detection), "All" represents the percentage of outputs that pass all criteria for that scenario, while "Avg" is the average of the three "All" scores across scenarios. Additionally, in error detection scenarios, Det measures the rate of error identification, and Help assesses the helpfulness of error messages for incorrect inputs.

The experimental results in Table \ref{tab:model_evaluation} demonstrate that GTM models achieve significant advantages in generating high-quality tool responses. Specifically,
GTM-1.5B (Avg: 89.4\%) further surpasses stronger baselines such Qwen2.5-14B-Instruct (85.8\%), despite having fewer parameters. A notable pattern across all models is their strong performance in single-turn scenarios (high "All" scores), where correct API calling conventions can be learned without context management challenges.
However, all model's performance declines in multi-turn scenarios, where preserving contextual continuity becomes crucial, here, GTM models demonstrate superior consistency, with GTM-1.5B achieving a multi-turn "All" score of 86.7\%, significantly outperforming baselines of similar or larger scale. 

\begin{table}[h]
\centering
\caption{Average response time comparison of search APIs and GTM.}
\label{tab:api_latency}
\resizebox{\columnwidth}{!}{%
\begin{tabular}{l|c|c}
\toprule
API Service & Response Time (s) & Source \\
\midrule
SerpApi & 0.73 & \cite{serpapi2024} \\
BrightData & 3.48 & \cite{serpapi2024} \\
DataForSEO & 3.57 & \cite{serpapi2024} \\
Scrapingdog & 0.89 & \cite{serpapi2024} \\
ScraperAPI & 14.84 & \cite{serpapi2024} \\
Serper & 0.83 & \cite{serpapi2024} \\
\midrule
Jina & 0.923 & Measured \\
\midrule
GTM (1×A800 40G) & 0.48 & Measured \\
GTM (2×A800 40G) & 0.46 & Measured \\
GTM (4×A800 40G) & 0.48 & Measured \\
\bottomrule
\end{tabular}%
}
\end{table}

\begin{figure}[htbp]
    \centering
    \includegraphics[width=\columnwidth]{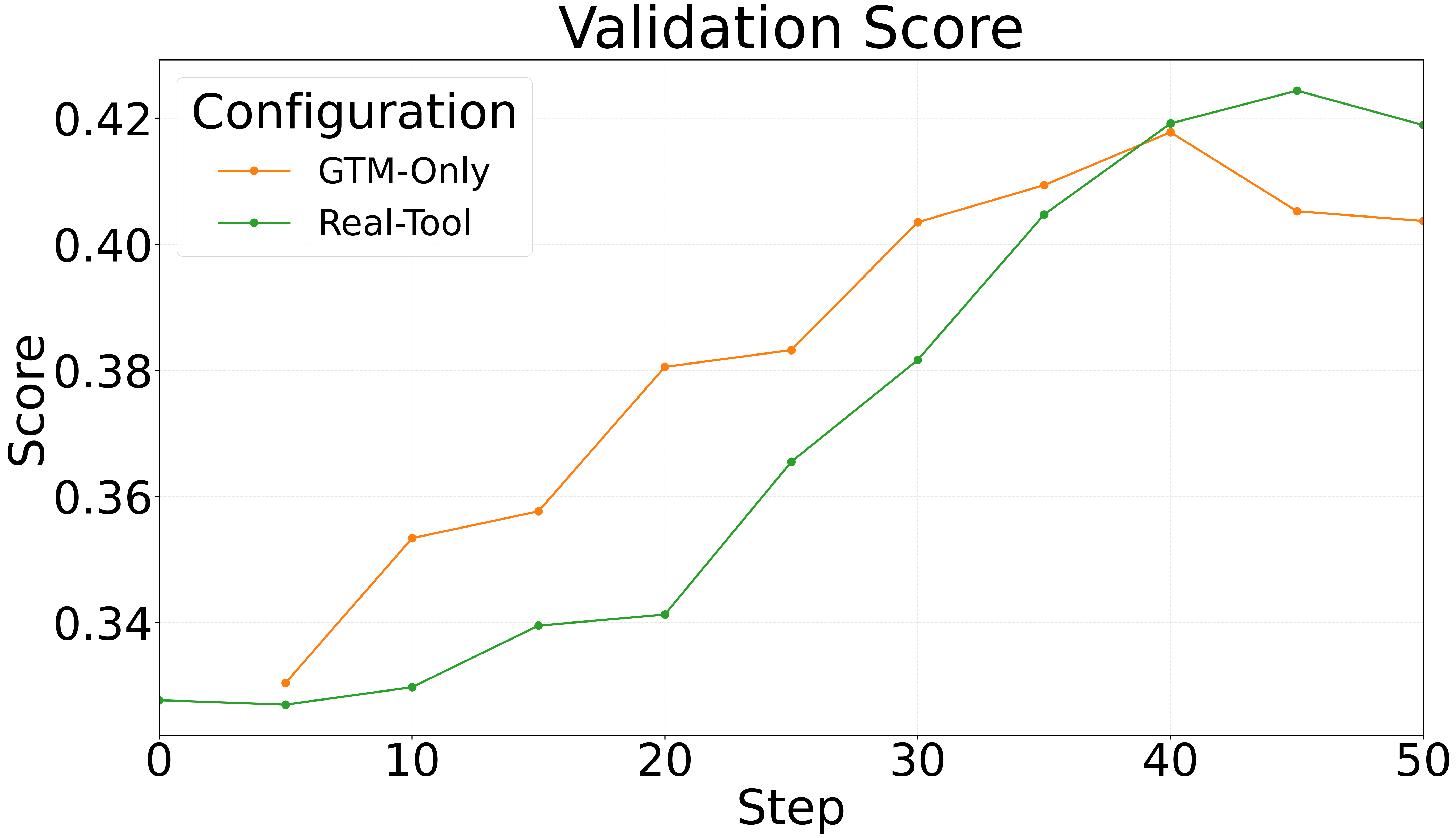}
    \caption{Validation scores in search tool scenarios. GTM-Only achieves 0.417 compared to Real-Tool's 0.424.}
    \label{fig:search_accuracy}
\end{figure}


\begin{figure}[htbp]
    \centering
    \includegraphics[width=\columnwidth]{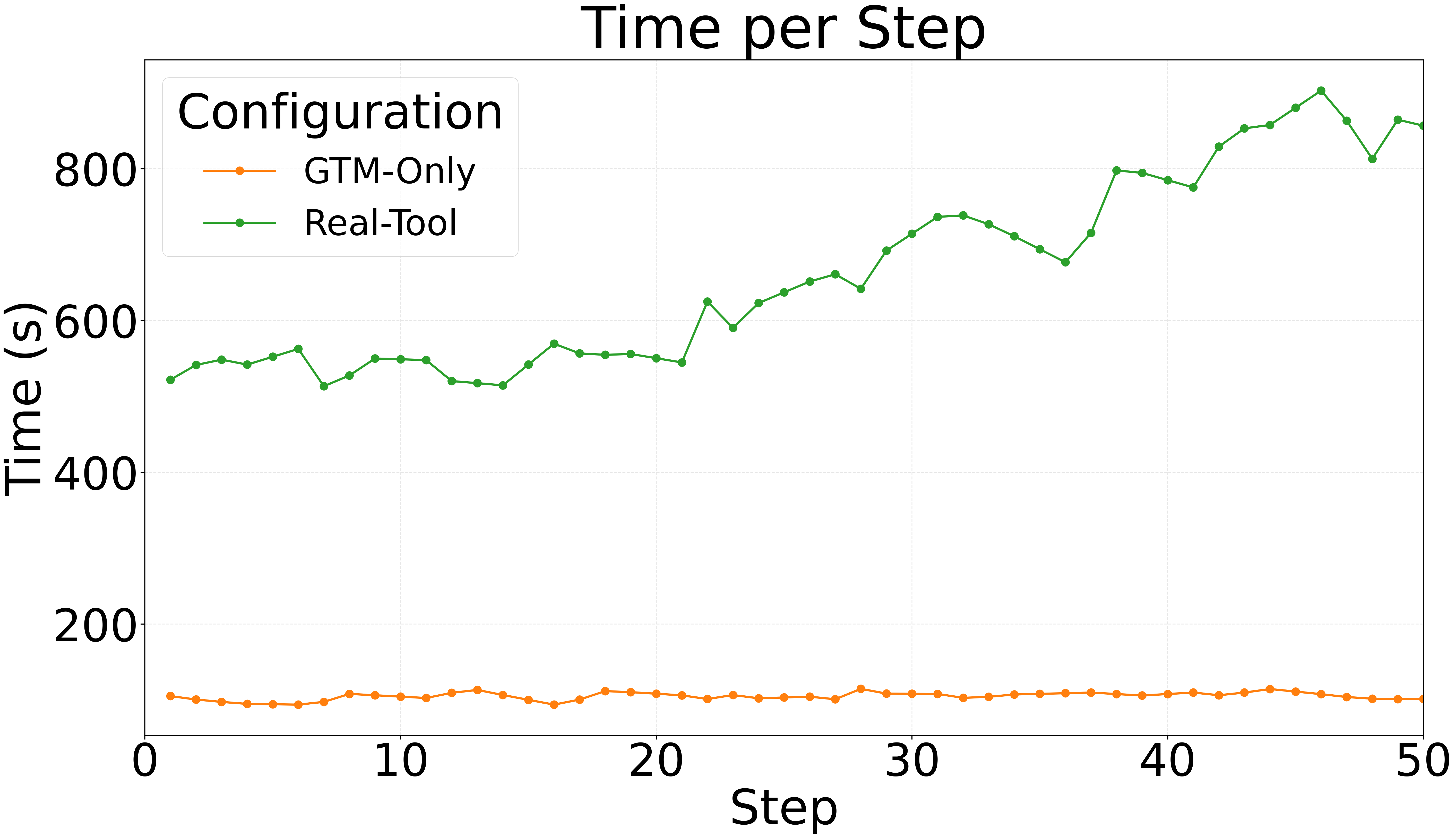}
    \caption{Average time per step in search tool scenarios. GTM-Only takes 5,255 seconds total compared to Real-Tool's 33,092 seconds, achieving 6× speedup.}
    \label{fig:search_time}
\end{figure}

\begin{figure}[htbp]
    \centering
    \includegraphics[width=\columnwidth]{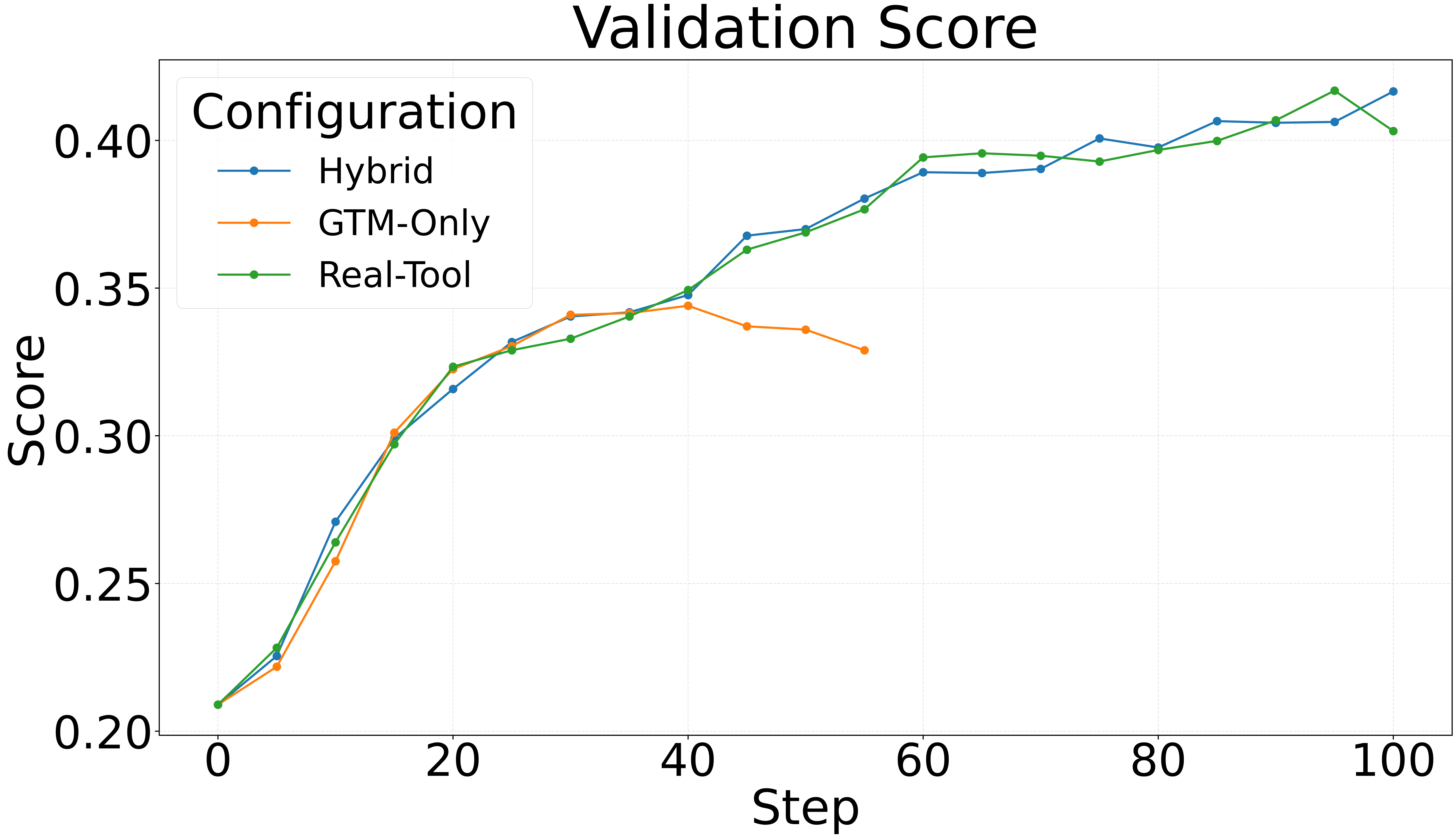}
    \caption{Evaluation scores in retrieval tool scenarios. GTM-Only achieves 0.341, while both Real-Tool and Hybrid reach 0.417.}
    \label{fig:retrieval_accuracy}
\end{figure}

\subsection{Use GTM to simulate tool}

We conducted experiments comparing two training configurations: GTM-Only, where the entire training process uses GTM-simulated tools, and Real-Tool, where training employs actual Jina search API throughout. Table \ref{tab:api_latency} compares the average response times of various search APIs. While commercial search APIs typically require 0.73-14.84 seconds per query \cite{serpapi2024}, with our measured Jina API averaging 0.923 seconds, GTM achieves consistent sub-second performance at 0.48 seconds regardless of GPU configuration. 

As shown in Figure~\ref{fig:search_accuracy}, both configurations started with an identical validation score of 0.328. After 40 training steps, GTM-Only reached its peak performance of 0.418, while Real-Tool achieved a slightly higher peak of 0.424 after 45 steps. Notably, although GTM's final performance was marginally lower than the real tool configuration, it demonstrated significantly faster convergence. By step 20, GTM-Only had already achieved a validation score of 0.381, while Real-Tool lagged behind at 0.341, indicating GTM's ability to accelerate early-stage learning.

Further analysis revealed substantial differences in efficiency metrics between the two approaches. As illustrated in Figure~\ref{fig:search_time}, the training time per step showed a dramatic difference: GTM required only 105.1 seconds per step on average, while Real-Tool consumed 661 seconds, making GTM 6.3× faster. These results demonstrate that while GTM achieves slightly lower final performance (approximately 1.4\% difference), it offers exceptional training efficiency, making it highly practical for rapid prototyping and development iterations where training speed is crucial.





\subsection{Use GTM to simulate unseen tool}

\begin{table*}[htbp]
\centering
\caption{Performance Comparison under Kernel bench level 1 task.}
\label{tab:Kernel_p}
\resizebox{0.8\textwidth}{!}{
\begin{tabular}{lccccc}
\toprule
\multirow{2}{*}{Speedup Threshold (p)} & \multirow{2}{*}{qwen2.5-7b}& qwen2.5-7b & qwen2.5-7b & \multirow{2}{*}{llama3.1-405b} & \multirow{2}{*}{qwen2.5-72b} \\
& & (real-40step) & (GTM-40step) & & \\
\midrule
1.0 & 0\% & 8\% & 5\% & 2\% & 6\% \\
1.5 & 0\% & 2\% & 2\% & 2\% & 3\% \\
2.0 & 0\% &  1\% & 1\% & 2\% & 2\% \\
\midrule
Geometric Mean of Speedup &\multirow{2}{*}{22.7\%}& \multirow{2}{*}{99.2\%} & \multirow{2}{*}{100.1\%} & \multirow{2}{*}{11.2\%} & \multirow{2}{*}{20.1\%}\\
for Correct Samples\\
\bottomrule
\end{tabular}%
}
\end{table*}

\begin{table}[htbp]
\centering
\caption{Performance Metrics Summary for Three Prediction Tasks.}
\resizebox{\columnwidth}{!}{%
\begin{tabular}{lccc}
\toprule
Metric & Compiled & Correctness & Runtime \\
\midrule
Accuracy & 90.65\% & 96.02\% & 99.45\%$^*$ \\
Precision & 93.65\% & 94.64\% & -- \\
Recall & 87.28\% & 71.14\% & -- \\
F1-Score & 90.35\% & 81.23\% & -- \\
\midrule
\multicolumn{4}{l}{Runtime-specific Metrics} \\
\midrule
Sample count (runtime $\neq$ -1.0) & -- & -- & 298 \\
Mean error & -- & -- & 7.04 ms \\
Median error & -- & -- & 0.10 ms \\
\bottomrule
\multicolumn{4}{l}{\small $^*$ Accuracy when runtime = -1.0 (2,151 samples correctly predicted)}
\end{tabular}%
}
\label{tab:performance_summary}
\end{table}

To evaluate GTM's generalization capability on unseen tools, we conduct experiments using a retrieval tool that was deliberately excluded from GTM's training set. Specifically, we removed all tools with description similarity greater than 0.7 to the retrieval tool to ensure zero exposure during training.
We train Qwen2.5-3B agents on information-seeking tasks using reinforcement learning under three configurations:
(1) \textit{Real-Tool}: using the actual retrieval tool throughout 100 training steps; 
(2) \textit{GTM-Only}: replacing all tool calls with GTM simulations; 
and (3) \textit{Hybrid}: using GTM for the first 30 steps as warm-up, then switching to real tools for the remaining 70 steps. This setup allows us to assess both GTM's simulation quality on unseen tools and its practical value in agent training.

Figure~\ref{fig:retrieval_accuracy} shows the accuracy curves for all three approaches. The Real-Tool baseline improves from 0.21 to 0.40 over 100 steps. GTM-Only achieves promising early results, reaching 0.34 accuracy by step 40, but subsequently collapses due to error accumulation. Notably, GTM's simulation quality remains high during the first 30 steps, motivating our Hybrid approach. This strategy leverages GTM's early-stage effectiveness while avoiding long-term degradation, achieving a final accuracy of 0.41, matching or slightly exceeding the Real-Tool baseline. These results demonstrate that GTM can effectively serve as a warm-up trainer for unseen tools without compromising final performance. We also compare the total time of the Hybrid method and Real-Tool method, which turns out that Hybrid method has a slightly speed up(about 15\%) during the warm up stage.

\subsection{Use GTM in Specific Domain}

\begin{figure}[htbp]
    \centering
    \includegraphics[width=\columnwidth]{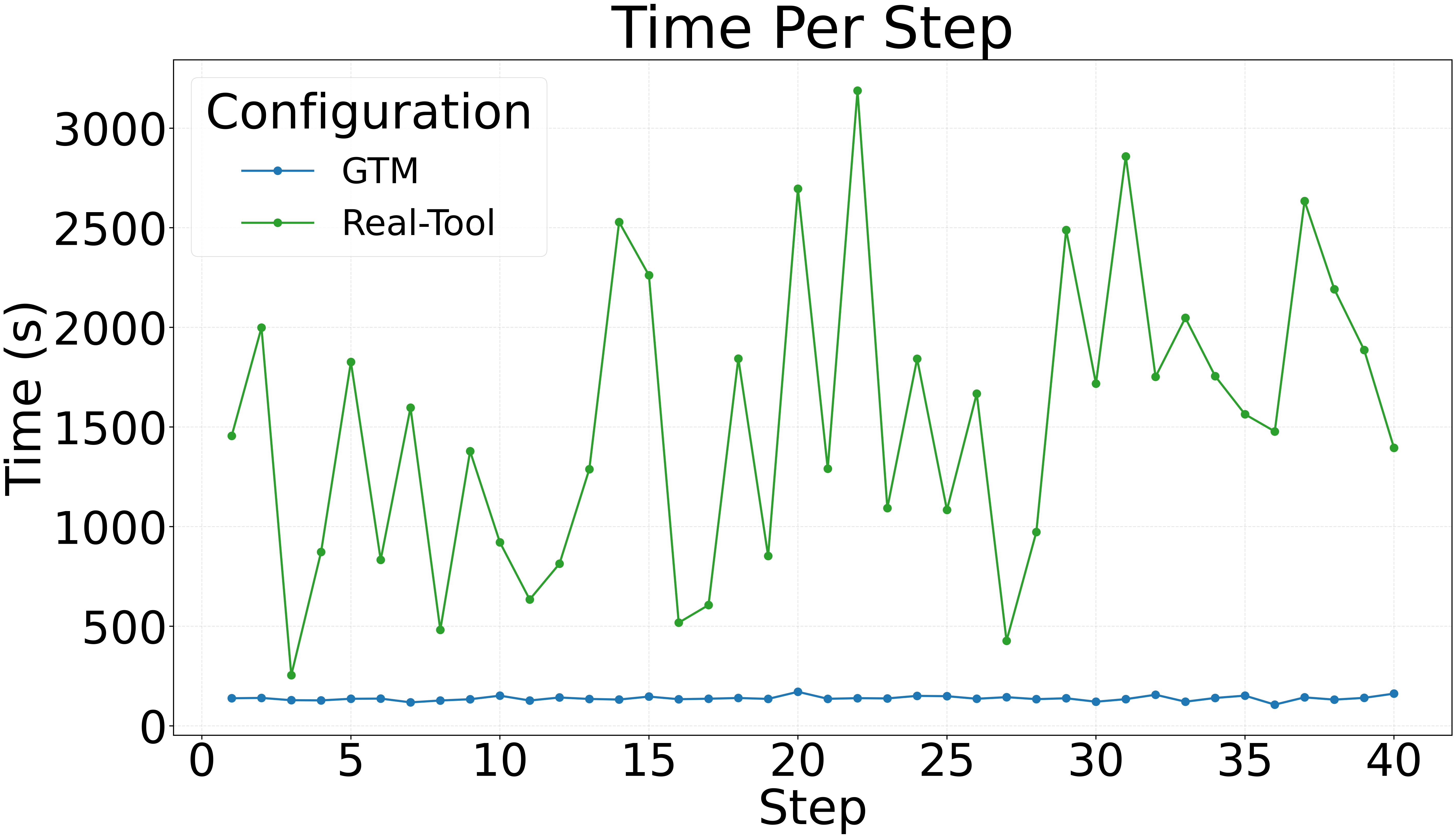}
    \caption{Time per step in CUDA kernel optimization scenarios. GTM-Only takes 5,518 seconds total compared to Real-Tool's 61,003 seconds, achieving 11× speedup.}
    \label{fig:cuda_time}
\end{figure}

While general-purpose GTM performs well across diverse tools, certain specialized domains require domain-specific knowledge for accurate simulation. We investigate GTM's adaptability through fine-tuning on CUDA kernel optimization, where agents learn to rewrite CUDA kernels for improved performance. The domain-specific tool, which we call it CUDA Code Validator, evaluates three metrics: compilation errors, runtime errors, and execution time by actually compiling and running the generated code. 
We collect training data from KernelBench, including various kernel implementations and their corresponding evaluation results, to fine-tune GTM for this specialized domain. For the agent training, we use Qwen2.5-7B as the base model. 
Our experiments examine: (1) the fine-tuned GTM's prediction accuracy, (2)final agent performance on KernelBench, and (3) training efficiency improvements.

The fine-tuned GTM accurately simulates the CUDA evaluation tool. As shown in Table \ref{tab:performance_summary}, GTM detects compilation errors with 90.65\% accuracy, predicts runtime errors with 99.45\% accuracy, and estimates execution time within 1 millisecond in 50\% of cases. This accuracy makes GTM a viable substitute for the actual tool during training.

Table \ref{tab:Kernel_p} shows GTM's performance as a CUDA Code Validator. The Speedup Threshold (p) measures performance relative to native CUDA, where p=1.0 indicates equivalent performance, p=1.5 shows 50\% improvement, and p=2.0 represents doubled speed. We use the geometric mean of speedup for correct samples to measure overall performance, as it reduces the influence of outliers. Although the GTM-trained model performs slightly below the real-tool trained version, it achieves results comparable to llama3.1-405b, confirming that GTM can effectively replace expensive real-tool training while maintaining competitive performance.

The efficiency gains are substantial. As shown in Figure \ref{fig:cuda_time}, training time drops from 61,003 seconds with the real tool to 5,518 seconds with GTM, an 11-fold speedup achieved by eliminating costly compilation and execution cycles. While agents trained with GTM show slightly lower performance than those trained with real tools, this trade-off is worthwhile given the dramatic reduction in training costs. 
This makes GTM a practical solution for developing domain-specific agents with limited computational resources.

\begin{figure*}[htbp]
    \centering
    \includegraphics[width=\textwidth]{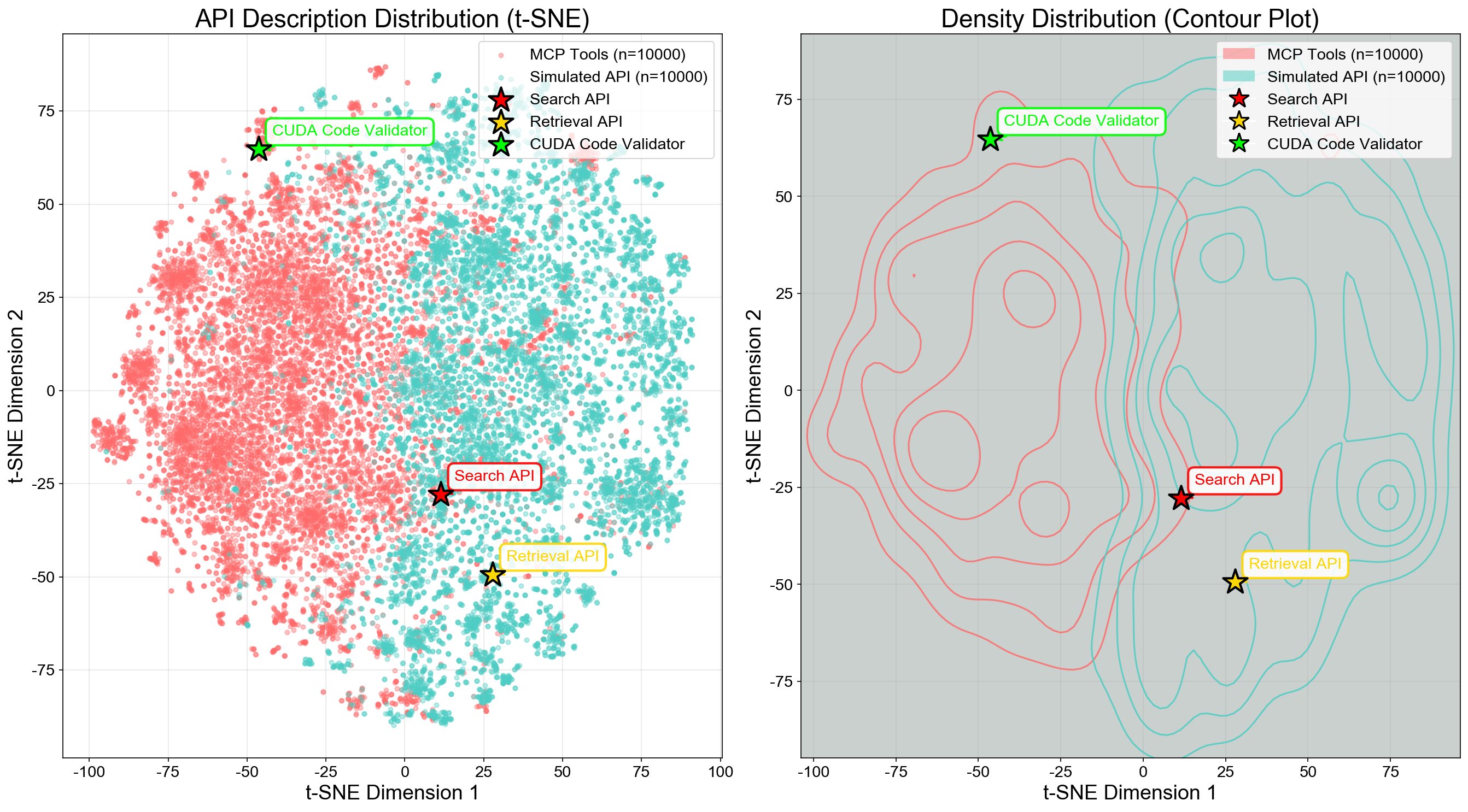}
    \caption{The distribution of real tool and our generated tool dataset.}
    \label{fig:dist_tool}
\end{figure*}

\subsection{Analysis application scope of GTM}

In this section, we conduct a boundary analysis to understand GTM's capabilities, specifically identifying scenarios where GTM can effectively replace real tools and situations where it cannot. To establish a representative benchmark of real-world tools, we leveraged the MCP (Model Context Protocol) tool marketplace at https://mcpmarket.cn. MCP is an open protocol that enables seamless integration between AI assistants and external tools, providing standardized interfaces for various functionalities ranging from file operations to API integrations. We crawled all MCP tools available on the platform as of December 1, 2025, obtaining over 39,000 real tool entries with their names and descriptions. Since the platform primarily serves Chinese users, we translated all MCP descriptions to English for consistent analysis. We then embedded both the real MCP tools and our simulated APIs used for training GTM using the paraphrase-multilingual-MiniLM-L12-v2 model and visualized the distribution using t-SNE.

As shown in Figure \ref{fig:dist_tool}, there is over 20\% overlap between MCP tools and simulated APIs, indicating substantial common ground in tool functionalities. Analysis of the non-overlapping regions reveals distinct patterns: MCP-exclusive tools predominantly include platform-specific integrations (e.g., Slack, Discord, GitHub APIs), server management utilities, and AI agent/model-related services—these represent tools tightly coupled to specific external services that are inherently difficult to simulate without access to the underlying systems. Conversely, simulated API-exclusive regions primarily contain database query operations, user-specific actions, and private domain operations that, while important for training, are unlikely to be exposed as public MCP tools due to privacy and security concerns. 

We also mapped our three experimental tools, search, retrieval, and CUDA code evaluator, onto this distribution. The search tool falls within the intersection of simulated APIs and MCP tools, explaining why GTM could successfully replace the real search API after absorbing substantial knowledge from similar tools in the training data. The retrieval tool, while centered in the simulated API region, lies distant from real MCP tools, suggesting that although GTM can simulate retrieval behavior based on learned patterns, it lacks exposure to real retrieval tool characteristics and thus cannot fully substitute actual retrieval systems. The CUDA code evaluator resides among real MCP tools but far from simulated APIs, indicating a knowledge gap that necessitates domain-specific fine-tuning for GTM to effectively simulate such specialized tools. This analysis provides clear guidance on GTM's applicability: it excels at simulating tools with substantial representation in training data but requires additional adaptation for highly specialized or platform-dependent tools.


\begin{table}[h]
\centering
\caption{Performance comparison of different base models before and after training with CARG-generated data. All values are percentages.}
\label{tab:ablation_models}
\resizebox{\columnwidth}{!}{%
\begin{tabular}{l|c|c|c|c}
\toprule
Model & Single-turn & Multi-turn & Error Detection & Avg \\
\midrule
Llama-3.2-1B-Instruct & 45.6 & 13.4 & 20.6 & 26.5 \\
\textbf{+ CARG} & \textbf{93.7} & \textbf{81.4} & \textbf{95.3} & \textbf{90.1} \\
\midrule
InternLM2.5-1.8B & 3.0 & 5.2 & 24.2 & 10.8 \\
\textbf{+ CARG} & \textbf{88.9} & \textbf{74.1} & \textbf{94.9} & \textbf{85.9} \\
\bottomrule
\end{tabular}%
}
\end{table}

\begin{table}[h]
\centering
\caption{Ablation study on multi-turn training data. GTM w/o multi-turn is trained only on single-turn and error generation data.}
\label{tab:ablation_multiturn}
\resizebox{\columnwidth}{!}{
\begin{tabular}{l|cccccc}
\toprule
Model & Format & Logic & Sem & Comp & Cons & All \\
\midrule
Base 1.5B & 74.1 & 74.5 & 68.9 & 84.8 & 98.4 & 53.0 \\
GTM w/o multi-turn & 95.3 & 89.2 & 91.5 & 96.5 & 99.0 & 83.0 \\
GTM (full) & 97.2 & 90.1 & 94.0 & 98.5 & 99.0 & 86.7 \\
\bottomrule
\end{tabular}
}

\end{table}

\subsection{Ablation Study}

To validate the effectiveness of our proposed CARG pipeline and its core design choices, we conducted two ablation experiments.

To validate the effectiveness of our proposed CARG pipeline, we applied it to train different base models. Table \ref{tab:ablation_models} shows the performance of Llama-3.2-1B and InternLM2.5-1.8B before and after training with CARG-generated data. Both models demonstrate dramatic improvements: Llama-3.2-1B improved from 26.5\% to 90.1\% average across all tasks, while InternLM2.5-1.8B showed even more remarkable gains from 10.8\% to 85.9\%. The most striking improvements occur in multi-turn scenarios where Llama-3.2-1B increased from 13.4\% to 81.4\% and InternLM2.5-1.8B from 5.2\% to 74.1\%, demonstrating CARG's effectiveness in teaching contextual consistency. These consistent improvements across different architectures validate that CARG's benefits stem from the quality of generated training data rather than model-specific factors.

To verify whether CARG's generation process can inherently teach models contextual awareness, we trained a variant of GTM-1.5B using only single-turn and error generation data, excluding all multi-turn dialogues from the training set. As shown in Table \ref{tab:ablation_multiturn}, while this single-turn-only variant improved significantly over the base model (83.0\% vs. 53.0\% on multi-turn ``All'' score), it still falls short of the full GTM's 86.7\%. These results suggest that CARG can enable models to acquire contextual awareness even from single-turn data alone. However, to achieve optimal multi-turn performance, incorporating dedicated multi-turn training data remains essential.

\section{Conclusion}

We presented the Generalist Tool Model (GTM), a universal tool simulator that enables efficient and scalable training of tool-augmented LLM agents by decoupling the learning process from expensive, slow, and unstable real tool interactions. Through our Context Awareness Response Generation (CARG) pipeline, GTM learns from over 20,000 tools spanning 300 domains, acquiring not only format correctness in API calling conventions, but also logical coherence in input-output relationships and contextual consistency across multi-turn interactions. Our extensive experiments demonstrate that GTM achieves simulation speeds orders of magnitude faster than real APIs while maintaining response quality comparable to actual tools, exhibiting strong generalization to unseen tools and effective domain adaptation through fine-tuning. By transforming the paradigm of tool-augmented agent training from direct API interaction to efficient simulation, GTM serves as a foundational component in agentic RL process.

\nocite{langley00}

\bibliography{example_paper}
\bibliographystyle{icml2025}


    



\end{document}